\documentclass[conference]{IEEEtran}

\usepackage{microtype}                  % pretty margins
\usepackage{enumitem}                   % margin-free itemize
\usepackage[normalem]{ulem}             % underline
\usepackage[hang,flushmargin]{footmisc} % footnote indentation
\usepackage{bm} % bold math symbols
\usepackage{times}
\usepackage{url}
\usepackage{latexsym}
\usepackage{fixltx2e}
\usepackage{graphicx}
\usepackage{tablefootnote}
\usepackage{algorithm2e}
\usepackage{multirow}
\usepackage{amsmath}
\usepackage{amssymb}
\usepackage{wasysym}
\usepackage{nicefrac}

\ifCLASSINFOpdf
\else
\fi

\begin{document}

\title{SelQA: A New Benchmark for Selection-based Question Answering}

\author{\IEEEauthorblockN{Tomasz Jurczyk}
\IEEEauthorblockA{Mathematics and Computer Science\\
Emory University\\
Atlanta, GA 30322, USA\\
\texttt{tomasz.jurczyk@emory.edu}}
\and
\IEEEauthorblockN{Michael Zhai}
\IEEEauthorblockA{Mathematics and Computer Science\\
Emory University\\
Atlanta, GA 30322, USA\\
\texttt{michael.zhai@emory.edu}}
\and
\IEEEauthorblockN{Jinho D.\ Choi}
\IEEEauthorblockA{Mathematics and Computer Science\\
Emory University\\
Atlanta, GA 30322, USA\\
\texttt{jinho.choi@emory.edu}}}

\maketitle

%Tomasz Jurczyk\qquad Michael Zhai\qquad Jinho D. Choi\\Mathematics and Computer Science\\
%            Emory University\\
%            Atlanta, GA 30322, USA\\
%            \texttt{\{tomasz.jurczyk,michael.zhai,jinho.choi\}@emory.edu

\begin{abstract}
This paper presents a new selection-based question answering dataset, SelQA. The dataset consists of questions generated through crowdsourcing and sentence length answers that are drawn from the ten most prevalent topics in the English Wikipedia.
We introduce a corpus annotation scheme that enhances the generation of large, diverse, and challenging datasets by explictly aiming to reduce word co-occurrences between the question and answers.
Our annotation scheme is composed of a series of crowdsourcing tasks with a view to more effectively utilize crowdsourcing in the creation of question answering datasets in various domains.
Several systems are compared on the tasks of answer sentence selection and answer triggering, providing strong baseline results for future work to improve upon.
\end{abstract}

%This paper presents a new dataset to benchmark selection-based question answering.
%Our dataset consists of contexts drawn from the top most prevalent topics in the English Wikipedia.
%We propose a new annotation scheme for the generation of a large, diverse, and challenging question answering and answer triggering dataset. Our anntation scheme involves a series of crowdsourcing tasks that % expand on annotations here
%We compare several systems on the task of answer sentence selection and answer triggering and provide strong baseline results for future work to improve upon.
%We hope that by providing a large answer sentence selection and triggering corpus will enable researchers to work towards more effective large scale open-domain question answering systems.

\IEEEpeerreviewmaketitle

\section{Introduction}

\noindent Selection-based question answering is the task of selecting a segment of text, or interchangeably a context, from a provided set of contexts that best answers a posed question.
Let us define a context, as a single document section, a group of contiguous sentences, or a single sentence.
Selection-based question answering is subdivided into answer sentence selection and answer triggering.
Answer sentence selection is defined as ranking sentences that answer a  question higher than the irrelevant sentences where there is at least a single sentence that answers the question in a provided set of candidate sentences.
Answer triggering is defined as selecting any number $(n>=0)$ of sentences from a set of candidate sentences that answers a question where the set of candidate sentences may or may not contain sentences that answer the question.
Several corpora have been created for these tasks~\cite{wang:07a,yang:15a,feng:15a}, allowing researchers to build effective question answering systems~\cite{yu:14a,wang:15a,severyn:15a} with the aim of improving reading comprehension through understanding and reasoning of natural language.
However, most of these datasets are constrained in the number of examples and scope of topics.
We attempt to mitigate these limitations to allow for a more through reading comprehension evaluation of open-domain question answering systems.

This paper presents a new corpus with annotated question answering examples of various topics drawn from Wikipedia.
An effective annotation scheme is proposed to create a large corpus that is both challenging and realistic.
Questions are additionally annotated with its topic, type, and paraphrase that enable comprehensive analyses of system performance on the answer sentence selection and answer triggering tasks.
%\footnote{question types $=: \{$what, who, when, how where, why, misc$\}$}
Two recent state-of-the-art systems based on convolutional and recurrent neural networks are implemented to analyze this corpus and to provide strong baseline measures for future work.
In addition, our systems are evaluated on another dataset, WikiQA~\cite{yang:15a}, for a fair comparison to previous work.
Our analysis suggests extensive ways of evaluating selection-based question answering, providing meaningful benchmarks to question answering systems.
The contributions of this work include:\footnote{All our work will be publicly available on GitHub.}

\begin{itemize}
\item Creating a new corpus for answer sentence selection and answer triggering (Section~\ref{sec:corpus}).
\item Developing QA systems using the latest advances in neural networks (Section~\ref{sec:systems}). 
\item Analyzing various aspects of selection-based question answering (Section~\ref{sec:experiments}).
\end{itemize}

\begin{figure*}[htbp!]
\centering
%\captionsetup{justification=centering}
\includegraphics[scale=0.62]{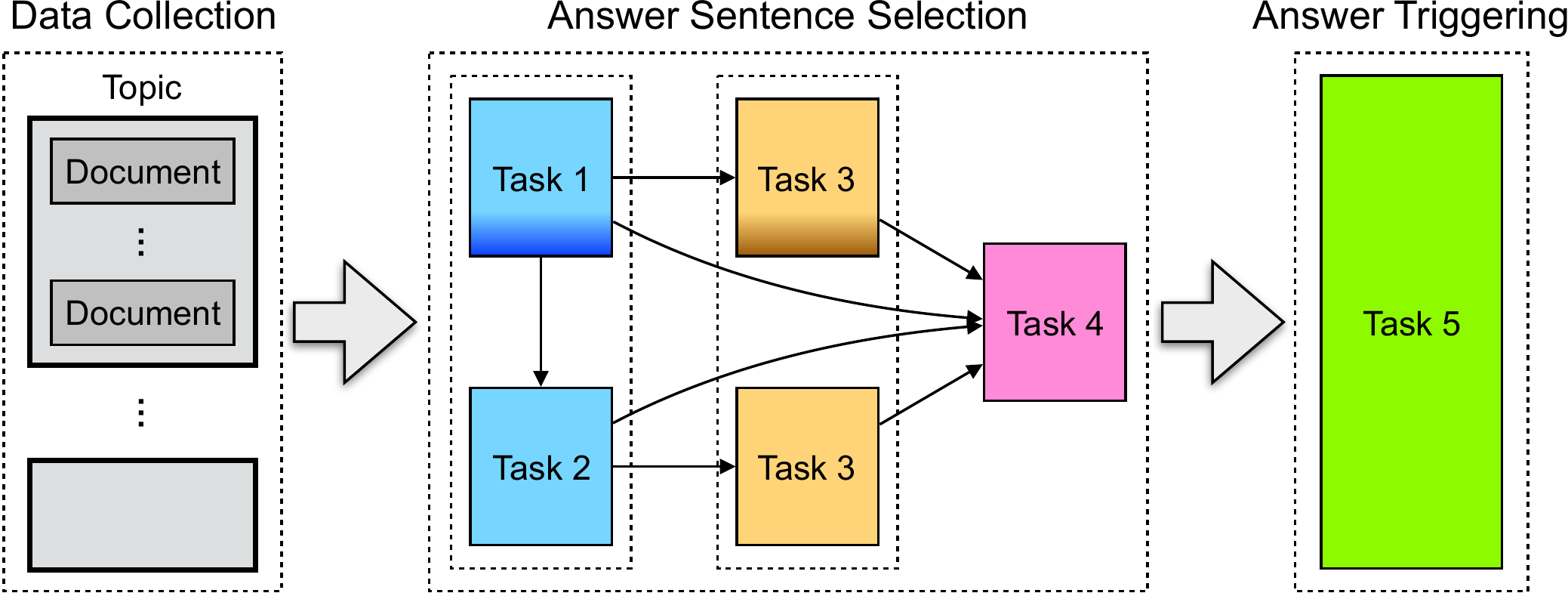}
\caption{The overview of our data collection (Section~\ref{ssec:collection}) and annotation scheme (Section~\ref{ssec:annotation}).}
\label{fig:annotation}
\end{figure*}

\section{Related Work}
\label{sec:related}

\noindent The TREC QA competition datasets have been a popular choice for evaluating answer sentence selection.\footnote{\url{http://trec.nist.gov/data/qa.html}} \cite{wang:07a} combined the TREC-[8-12] datasets for training and divided the TREC-13 dataset for development and evaluation.
This dataset, known as QASent, has been used as the standard benchmark for answer sentence selection although it is rather small (277 questions with manually picked answer contexts).
\cite{yang:15a} introduced a lager dataset, WikiQA, consisting of questions collected from the user logs of the Bing search engine.
Our corpus is similar to WikiQA but covers more diverse topics, consists of a larger number of questions (about 6 times larger for answer sentence selection and 2.5 times larger for answer triggering), and makes use of more contexts by extracting contexts from the entire article instead of from only the abstract.
\cite{feng:15a} distributed another dataset, InsuranceQA, including questions in the insurance domain.
WikiQA introduced the task of answer triggering and was the only answer triggering dataset.
Our corpus provides a new automatically generated answer triggering dataset.
%  where the contexts are automatically retrieved by Elasticsearch\footnote{\url{www.elastic.co/products/elasticsearch}} that can be easily reproduced.

Due to increasing complexity in question answering, deep learning has become a popular trend in solving difficult problems. \cite{yu2014deep} proposed a convolutional neural network with a single convolution layer, average pooling and logistic regression at the end for factoid question answering. Further, more convolutional neural network based frameworks have been proposed as solutions for question answering~\cite{iyyer2014neural,dong2015question,yin2015abcnn,yih2014semantic,blunsom2014convolutional}
Our convolutional neural network model is inspired by the previous work utilizing the tree-edit distance and the tree kernel~\cite{heilman:10a,wang:10a,severyn:13a}, although we introduce a different way of performing subtree matching facilitating word embeddings.
Our recurrent neural network models with attention are based on established state-of-the-art systems for answer sentence selection~\cite{tan:15a,santos:16a}.

\section{Corpus}
\label{sec:corpus}

\noindent Our annotation scheme provides a framework for any researcher to create a large, diverse, pragmatic, and challenging dataset for answer sentence selection and answer triggering, while maintaining a low cost using crowdsourcing.

% =================================== Data Collection ===================================

\subsection{Data Collection}
\label{ssec:collection}

\noindent A total of 486 articles are uniformly sampled from the following 10 topics of the English Wikipedia, dumped on August, 2014:

\begin{center}
\textit{Arts}, \textit{Country}, \textit{Food}, \textit{Historical Events},\\ \textit{Movies}, \textit{Music}, \textit{Science}, \textit{Sports}, \textit{Travel}, \textit{TV}.
\end{center}

\noindent These are the most prevalent topics categorized by DBPedia.\footnote{\url{http://dbpedia.org}}
The original data is preprocessed into smaller chunks.
First, each article is divided into sections using the section boundaries provided in the original dump.\footnote{\url{https://dumps.wikimedia.org/enwiki}}
Each section is segmented into sentences by the open-source toolkit, NLP4J.\footnote{\url{https://github.com/emorynlp/nlp4j}}
In our corpus, documents refer to individual sections in the Wikipedia articles.

\begin{table}[htp!]
\caption{Lexical statistics of our corpus.}
\label{tbl:annotation:wiki}
\centering
\begin{tabular}{l||r}
\multicolumn{1}{c||}{\textbf{Type}} & \multicolumn{1}{c}{\textbf{Count}} \\
\hline\hline
Total \# of articles  &       486 \\
Total \# of sections  &     8,481 \\
Total \# of sentences &   113,709 \\
Total \# of tokens    & 2,810,228 \\
\end{tabular}
\end{table}

% =================================== Annotation Scheme ===================================

\subsection{Annotation Scheme}
\label{ssec:annotation}

\noindent Four annotation tasks are conducted in sequence on Amazon Mechanical Turk for answer sentence selection (Tasks 1-4), and a single task is conducted for answer triggering using only Elasticsearch (Task 5; see Figure~\ref{fig:annotation} for the overview).
%Figure~\ref{fig:annotation:fig1} shows the overview of our annotation scheme.

\begin{table*}[ht]
\caption{Given a section, Task~1 asks to generate a question regarding to the section. Task~2 crosses out the sentence(s) related to the first question (line~1), and asks to generate another question. Task~3 asks to paraphrase the first two questions. Finally, Task~4 asks to rephrase ambiguous questions.}
\label{tbl:annotation:example}
\centering
\begin{tabular}{|ll|}
\hline
\multicolumn{2}{|c|}{\textbf{Topic}: TV, \textbf{Article}: Criminal Minds, \textbf{Section}: Critical reception}\\
& \\
\multicolumn{2}{|l|}{1. \uwave{The premiere episode was met with mixed reviews, receiving a score of 42 out of 100 on aggregate review site }}\\
\multicolumn{2}{|l|}{$\:\:\:\:\:$\uwave{Metacritic, indicating ``mixed or average'' reviews.}}\\
\multicolumn{2}{|l|}{2. Dorothy Rabinowitz said, in her review for the Wall Street Journal, that ``From the evidence of the first few episodes,}\\
\multicolumn{2}{|l|}{$\:\:\:\:\:$\textit{Criminal Minds} may be a hit, and deservedly''...}\\% and gave particular praise to both Matthew Gray Gubler and Mandy Patinkin's performance.\\
\multicolumn{2}{|l|}{3. The New York Times was less than positive, saying ``The problem with \textit{Criminal Minds} is its many confusing maladies,}\\
\multicolumn{2}{|l|}{$\:\:\:\:\:$applied to too many characters'' and felt that ``as a result, the cast seems like a spilled trunk of broken toys, with which}\\
\multicolumn{2}{|l|}{$\:\:\:\:\:$the audience - and perhaps the creators - may quickly become bored.''}\\
\multicolumn{2}{|l|}{4. The Chicago Tribune reviewer, Sid Smith, felt that the show ``May well be worth a look'' though he too criticized}\\
\multicolumn{2}{|l|}{$\:\:\:\:\:$``the confusing plots and characters''.}\\
%\multicolumn{2}{|c|}{$\ldots$}\\
%5. PopMatters panned the show, saying the show ``confuses critical thinking with supernatural abilities'' and criticized the\\``stereotypical characters''.\\
%6. The Los Angeles Times gave a similar review.\\
%7. However, both reviewers praised Patinkin and Gubler's performances.\\
\hline\hline
Task 1     & How was the premiere reviewed? \\
Task 2     & Who felt that Criminal Minds had confusing characters?\\
Task 3.1   & How were the initial reviews?\\
Task 3.2   & Who was confused by characters on Criminal Minds?\\
Task 4.3.1 & How were the initial reviews in Criminal Minds?\\
\hline
\end{tabular}
\vspace{-2ex}
\end{table*}

\begin{table*}[htbp!]
\caption{Q\textsubscript{s$|$m}: number of questions whose answer contexts consist of single$|$multiple sentences, $\Omega$\textsubscript{q$|$a}: macro avg.\ of overlapping words between $q$ and $a$, normalized by the length of $q|a$, $\Omega\textsubscript{f} = \nicefrac{(2\cdot\Omega\textsubscript{q}\cdot\Omega\textsubscript{a})}{(\Omega\textsubscript{q} + \Omega\textsubscript{a})}$, Time$|$Credit: avg.\ time$|$credit per mturk job. WikiQA statistics here discard questions w/o answer contexts.}
\label{tbl:annotation-stat}
\centering
%\resizebox{\columnwidth}{!}{
\begin{tabular}{c||r|r|r||c|c|c||c|c}
 & \multicolumn{1}{c|}{\textbf{Q\textsubscript{s}}} & \multicolumn{1}{c|}{\textbf{Q\textsubscript{m}}} & \multicolumn{1}{c||}{\textbf{Q\textsubscript{s+m}}} & \bm{$\Omega$\textsubscript{q}} & \bm{$\Omega$\textsubscript{a}} & \bm{$\Omega$\textsubscript{f}} & \bf Time & \bf Credit \\
\hline\hline
Task 1 & 1,824 & 154 & 1,978 & 44.99 & 23.65 & 28.88 & 71 sec. & \$ 0.10 \\
Task 2 & 1,828 & 148 & 1,976 & 44.64 & 23.20 & 28.62 & 64 sec. & \$ 0.10 \\
Task 3 & 3,637 & 313 & 3,950 & 38.03 & 19.99 & 24.41 & 41 sec. & \$ 0.08 \\
Task 4 &   682 &  55 &   737 & 31.09 & 19.41 & 21.88 & 54 sec. & \$ 0.08 \\
\hline\hline
Our corpus  & 7,289 & 615 & \bf 7,904 & 40.54 & 21.51 & \bf 26.18 & - & - \\
WikiQA      & 1,068 & 174 &     1,242 & 39.31 &  9.82 &     15.03 & - & - \\
\end{tabular}%}
\vspace{-2ex}
\end{table*}

\subsection*{Task 1}

\noindent Approximately two thousand sections are randomly selected from the 486 articles in Section~\ref{ssec:collection}.
All the selected sections consist of 3 to 25 sentences; we found that annotators experienced difficulties accurately and timely annotating longer sections.
For each section, annotators are instructed to generate a question that can be answered in one or more sentences in the provided section, and select the corresponding sentence or sentences that answer the question.
The annotators are provided with the instructions, the topic, the article title, the section title, and the list of numbered sentences in the section (Table~\ref{tbl:annotation:example}).

\subsection*{Task 2}

\noindent Annotators are asked to create another set of $\approx$2K questions from the same selected sections excluding the sentences selected as answers in Task~1.
The goal of Task~2 is to generate questions that can be answered from sentences different from those used to answer questions generated in the Task~1.
The annotators are provided with the same information as in Task~1, except that the sentences used as the answer contexts in Task~1 are crossed out (line 1 in Table~\ref{tbl:annotation:example}). Annotators are instructed not to use these sentences to generate new questions.

\subsection*{Task 3}

\noindent Although our instruction encourages the annotators to create questions in their own words, annotators will generate questions with some lexical overlap with the corresponding contexts.
The intention of this task is to mitigate the effects of annotators' tendency to  generating questions with similar vocabulary and phrasing to answer contexts. This is a necessary step in creating a corpus that evaluates reading comprehension rather than ability to model word co-occurrences.
%Annotators are asked to create a new set of questions by paraphrasing the questions generated from the previous tasks.
The annotators are provided with the previously generated questions and answer contexts and are instructed to paraphrase these questions using different terms.

\subsection*{Task 4}

\noindent Most questions generated by Tasks~1-3 are of high quality, that is they can be answered by a human when given the corresponding contexts; however, there are some questions that are ambiguous in meaning and difficult for humans to answer correctly.
These difficult questions often incorrectly assume that the related sections are provided with the questions. %, which cannot be assumed in reality.
For instance, it is impossible to answer the question from Task~3.1 in Table~\ref{tbl:annotation:example} unless the related section is provided with the question.
These ambiguous questions are sent back to the annotators for revision.

Elasticsearch is used to find ambiguous questions,\footnote{\url{www.elastic.co/products/elasticsearch}} a Lucene-based open-source search engine.
First, an inverted index of 8,481 sections is built, where each section is considered a document.
Each question is queried to this search engine.
If the answer context is not included within the top 5 sections in the search result, the question is considered `suspicious' although it may not be ambiguous.
Among 7,904 questions generated by Tasks~1-3, 1,338 of them are found to be suspicious.
These questions are sent to the annotators, and rephrased by the annotators if deemed necessary.

\subsection*{Task 5}
\label{ssec:answer-triggering-corpus}

\noindent By using the previously generated answer sentence selection data, the answer triggering corpus can be automatically generated again using Elasticsearch.
To generate answer contexts for answer triggering, all 14M sections from the entire English Wikipedia are indexed, and each question from Tasks 1-4 is queried.
Every sentence in the top 5 highest scoring sections from Elasticsearch are collected as candidates, which may or may not include the answer context that resolves the question.

%In our corpus, answer candidates are drawn from 1 section for answer sentence selection, whereas they are drawn from 5 sections for answer triggering.
%\noindent In this task, the entire question is taken as a query in Elasticsearch and 5 sections with the highest relevance scores are collected.
%We follow the default configuration of Elasticsearch and apply a simple tf-idf as a similarity measurement and relevance scoring. 
%stopwords are excluded from indexing, and BM25~\cite{robertson2004simple} is used as a similarity measurement.
%This Lucene approach gives an efficient way of generating high-quality annotation.
%Among the 1,338 suspicious questions, 55\% of them are paraphrased by the annotators.

% =================================== Analytics ===================================

\begin{figure*}[ht]
\centering
\includegraphics[scale=0.6]{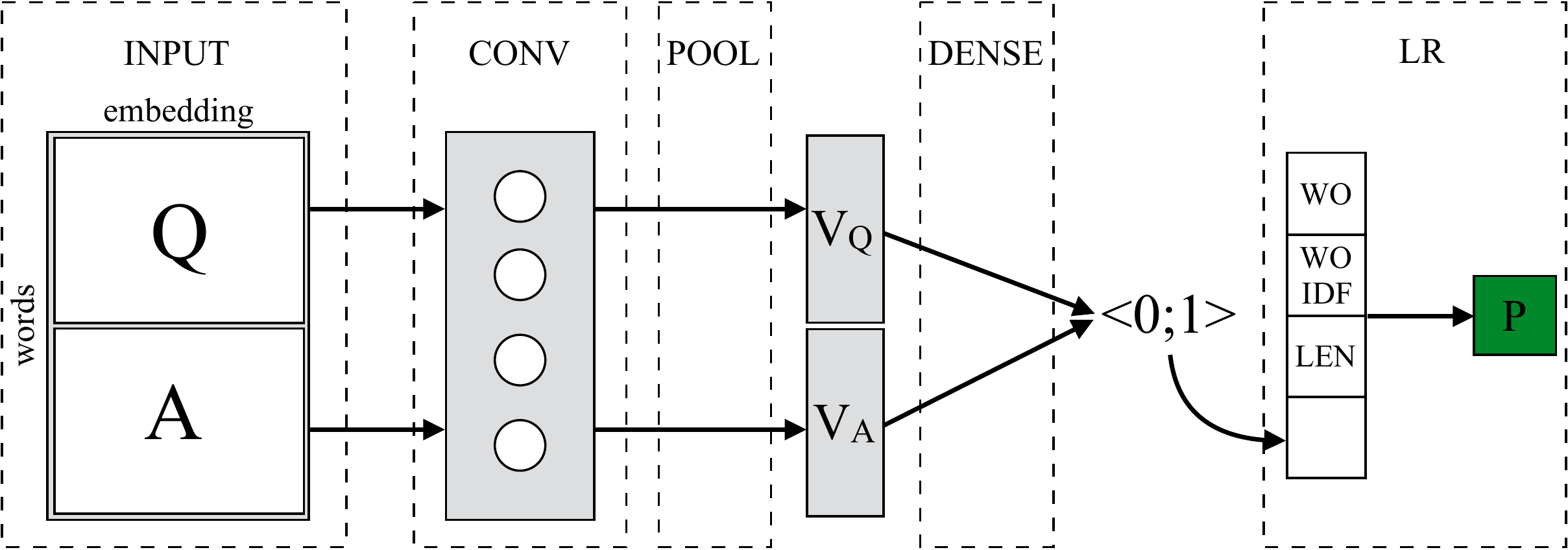}
\caption{The overview of our system using a convolutional neural network and logistic regression.}
\label{fig:fig1}
\vspace{-2ex}
\end{figure*}

\subsection{Corpus Analysis}
\label{ssec:corpus-analysis}

%Table~\ref{tbl:annotation-stat} shows the statistics of our corpus and annotation.
\noindent The entire annotation took about 130 hours, costing $\$770$ in total; each mturk job took on average approximately 1 minute and costed about $\cent 10$.
A total of 7,904 questions were generated from Tasks~1-4, where 92.2\% of them found their answers in single sentences.
It is clear that Task~3 was effective in reducing the percentage of overlapping words between question and answer pairs (about 4\%; $\Omega$\textsubscript{f} in Table~\ref{tbl:annotation-stat}).
The questions from Task 3 can be used to develop paraphrasing models as well.
Multiple pilot studies on different tasks were conducted to analyze quality and cost; Tasks 1-4 were proved to be the most effective in the pilot studies.
Following \cite{ho:15a}, we paid incentives to those who submitted outstanding work, which improved the overall quality of our annotation.
%\newcite{harris:11a}

Our corpus could be compared to WikiQA that was created with the intent of providing a challenging dataset for selection-based question answering~\cite{yang:15a}.
Questions in this dataset were collected from the user logs of the Bing search engine, and associated with the specific sections in Wikipedia, namely the first sections known as the abstracts.
We aim to provide a similar yet more exhaustive dataset by broadening the scope to all sections.
%Considering a larger context adds a layer of complexity for locating the correct section than considering only the abstracts.
A notable difference was found between these two corpora for overlapping words (about 11\% difference), which was expected due to the artificial question generation in our scheme.
Although questions taken from the search queries are more natural, real search queries are inacessible to most researchers.
The new annotation scheme proposed here can prove useful for researchers needing to create a corpus for selection-based QA.

Our answer triggering dataset contains 5 times more answer candidates per question than WikiQA because WikiQA includes only sections clicked on by users. Manual selection is eliminated from our framework, making our corpus more practical. % since finding the relevant section no longer rely on the user clicks.
In WikiQA, 40.76\% of the questions have corresponding answer contexts for answer triggering, as compared to 39.25\% in ours.

\section{Systems}
\label{sec:systems}

\noindent Two models using convolutional neural networks are developed, one is our replication of the best model in \cite{yang:15a}, and the other is an improved model using subtree matching (Section~\ref{ssec:cnn}).
Two more models using recurrent neural networks are developed, one is our replication of the attentive pooling model in \cite{santos:16a}, and the other is a simpler model using one-way attention (Section~\ref{ssec:rnn}).
These are inspired by the latest state-of-the-art approaches, providing sensible evaluations.

% =================================== Convolutional Neural Networks ===================================

\subsection{Convolutional Neural Networks}
\label{ssec:cnn}

\noindent Our CNN model is motivated by \cite{yang:15a}.
First, a convolutional layer is applied on the image of text using the hyperbolic tangent activation function.
The image consists of rows standing for consecutive words in two sentences, the question ($q$) and the answer candidate ($a$), where the words are represented by their embeddings~\cite{mikolov:13b}. For our experiments, we use the image of 80 rows (40 for question and answer, respectively). If any of the question or answer is longer than 40 tokens, the rest is being cut from the input.
%The image has 80 rows, 40 for $q$ and another 40 for $a$.
Next, the max pooling is applied,\footnote{We also experimented with the average pooling as \cite{yang:15a}, which led to a marginally lower accuracy.} and the sentence vectors for $q$ and $a$ are generated.
Unlike \cite{yang:15a} who performed the dot product between these two vectors, we added another hidden layer to learn their weights.
%Moreover, the word embeddings were retrained in our model, which was not done previously.
Finally, the sigmoid activation function is applied and the entire network is trained using the binary cross-entropy.

\noindent Next, we use a logistic regression model, where the CNN score from the output layer is used as one of the features.
Other features in the logistic regression are the number of overlapping words between $q$ and $a$, say $\Omega$, $\Omega$ normalized by the IDF, and the question length. While the logistic regression model could be merged directly with our CNN model, it has been empirically shown that it is more effective to construct this last phase as a separate model.

\begin{figure*}[ht]
\centering
%\captionsetup{justification=centering}
\includegraphics[scale=0.65]{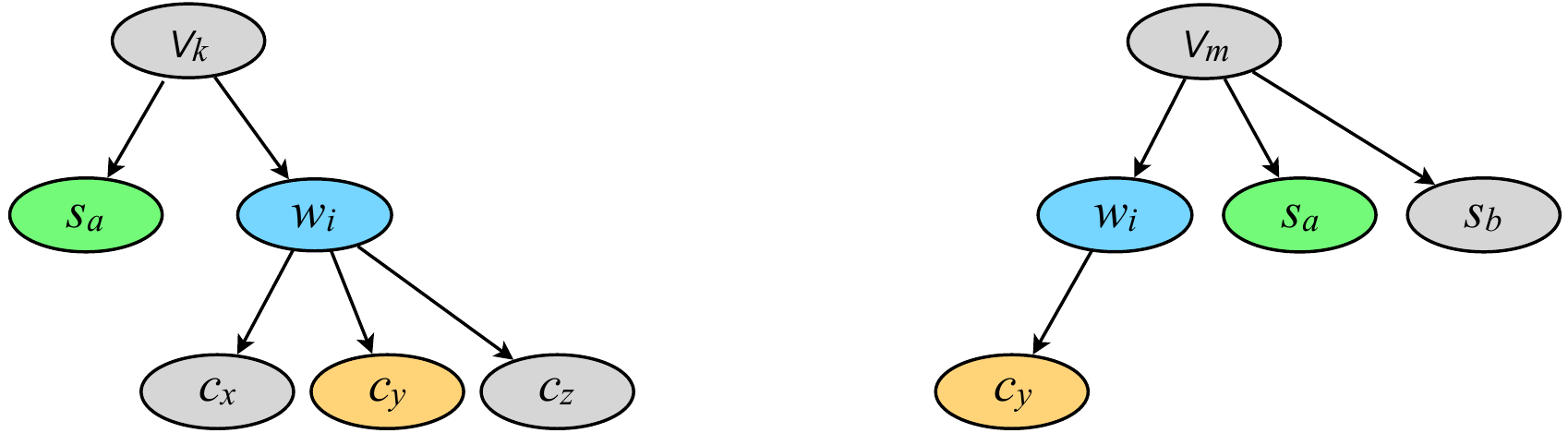}
\caption{Subtree matching between $D^q$ (left) and $D^a$ (right). $w_i$ is the $i$'th co-occurring word between $q$ and $a$. The color odes imply `match', and the grey nodes imply `non-match'. For instance, $v_k$ in $D^q$ is not matched to any node in $D^a$, whereas $c_y$ in $D^q$ finds its match in $D^a$.}
\label{fig:dependency-match}
\vspace{-2ex}
\end{figure*}

\begin{algorithm}[htp!]
  \KwIn{\small $T$: a set of co-occurring words between a question and answer.\linebreak $D^q,D^a$: sets of slices for a question and answer.\linebreak $f_m$: a metrics function.\linebreak $f_c$: a comparator function.}
  \KwOut{\small $S_{dep}$: A triplet of dependency similarity.}
   $S_{dep} \leftarrow$ $[0, 0, 0]$;\\
 
  \ForEach{\textrm{word} $w^o_i$ in $T$}{
    $p^q_i \leftarrow $getParent($D^q_i$)\\
    $p^a_i \leftarrow $getParent($D^a_i$)\\
    $S_{dep}[0] \leftarrow S_{dep}[0] + f_c(p^q_i, p^a_i)$\\
    $S^q_i \leftarrow $getSiblings($D^q_i$)\\
    $S^a_i \leftarrow $getSiblings($D^a_i$)\\
    $vals \leftarrow []$\\
    \ForEach{\textrm{sibling} $s^q_j$ in $S^q_i$}{
      \ForEach{\textrm{sibling} $s^a_k$ in $S^a_i$}{
        $vals$.append($f_c(s^q_j, s^a_k))$\\
      }
    }
    $S_{dep}[1] \leftarrow S_{dep}[1] + f_m(vals)$\\
    $C^q_i \leftarrow $getChildren($D^q_i$)\\
    $C^a_i \leftarrow $getChildren($D^a_i$)\\
    $vals \leftarrow []$\\
    \ForEach{\textrm{child} $c^q_j$ in $C^q_i$}{
      \ForEach{\textrm{child} $c^a_k$ in $C^a_i$}{
        $vals$.append($f_c(c^q_j, c^a_k))$\\
      }
    }
    $S_{dep}[2] \leftarrow S_{dep}[2] + f_m(vals)$\\
  }
 
 \caption{Algorithm of our subtree matching mechanism}
 \label{alg:dep-algorithm}
\end{algorithm}
\pagebreak

\noindent For the answer sentence selection task, the predictions for each question are treated as a ranking and the MAP and MRR scores are being calculated (Section~\ref{ssec:answer-sentence-selection}). On the other hand, in the answer triggering task (Section~\ref{ssec:answer-triggering}) a threshold is applied on each predicted question by the logistic regression; the candidate with the highest score is considered the answer if it is above the threshold found during development; otherwise, the model assumes no existence of the answer context in this document for that question. Figure~\ref{fig:fig1} shows the overview of our CNN and LR model.

% =================================== Subtree Matching ===================================

\subsubsection*{Subtree Matching} 
\label{sssec:subtree-matching}

\noindent We propose a subtree matching mechanism for measuring the contextual similarity between two sentences.
All sentences are automatically parsed by the NLP4J dependency parser~\cite{choi:13a}.
First, a set of co-occurring words between $q$ and $a$, say $T$, is created.
For each $w_i^o \in T$, $w_i^o$'s parents ($p_i^q$, $p_i^a$), siblings ($S_i^q$, $S_i^a$), and children ($C_i^q$, $C_i^a$) are extracted from the dependency slices of $q$ and $a$.
%\noindent Then, three matching scores are measured as follows:\footnote{We also experimented in presenting the subtree matching score as a single value and found that it lowered the accuracy.}
%\begin{align*}
%\mu_p &= \sum_{w_i \in T} f_c(p_i^q, p_i^a)\\
%\mu_s &= \sum_{w_i \in T} f_m(\{f_s(x, y) : \forall (x,y) \in S_i^q \times S_i^a\})\\
%\mu_c &= \sum_{w_i \in T} f_m(\{f_c(x, y) : \forall (x,y) \in C_i^q \times C_i^a\})
%\end{align*}
\noindent When the word-forms are used as the comparator, $f_c(x,y)$ returns $1$ if $x$ and $y$ have the same form; otherwise, $0$.
When the word embeddings are used as the comparator, $f_c(x,y)$ returns the cosine similarity between $x$ and $y$.
The function $f_m$ takes a list of scores and returns either the \texttt{sum}, \texttt{avg}, or \texttt{max} of the scores.
%Finally, $\mu_p$, $\mu_s$, and $\mu_c$ are used as the additional features to the logistic regression model.
Finally, the triplet $S_{dep}$ is used as the additional features to the logistic regression model.
Algorithm~\ref{alg:dep-algorithm} presents the entire process in detail.
Although our subtree matching mechanism adds only 3 more features, our experiments show significant performance gains for both the answer sentence selection and answer triggering strengthening our hypothesis that to solve question answering problems more effectively, deeper contextual similarity is required.

% =================================== Recurrent Neural Networks ===================================

\subsection{Recurrent Neural Network}
\label{ssec:rnn}

Our RNN model is based on the bidirectional Long Short-Term Memory (LSTM) using attentive pooling introduced by \cite{santos:16a}, except that our network uses a gated recurrent unit (GRU;~\cite{cho:14a}) instead of LSTM.
From our preliminary experiments, we found that GRU converged faster than LSTM while achieving similar performance for these tasks.
Let $w_i^q \in q$, $w_j^a \in a$, where $q$ is the question and $a$ is the answer candidate, and $e(w)$ returns the embedding of a word $w$.
Embeddings are encoded by a single bidirectional GRU $g$ that consists of the forward ($\overrightarrow{g}$) and the backward ($\overleftarrow{g}$) GRUs, each with $h$ hidden units.
Given $w$, $g$ outputs the vector concatenation of the hidden states of $\overrightarrow{g}$ and $\overleftarrow{g}$:
$$
g(e(w)) = \overrightarrow{g}(e(w)) || \overleftarrow{g}(e(w))
$$
Let $c = 2 \cdot h$ represent the dimensionality of the output of $g$.
Then, sentence embedding matrices $Q \in \mathbb{R}^{|q| \times c}$ and $A \in \mathbb{R}^{|a| \times c}$ are generated by $g$ as $Q_i = g(e(w^q_i))$ and $A_j = g(e(w^a_j))$.

Both the attentive pooling and one-way attention models below are trained by minimizing the pairwise hinge ranking loss.
In addition, RMSProp is used for the optimization and the $\ell_2$ weight penalty is applied on all parameters except for embeddings.
All network parameters except the embeddings are initialized using orthogonal initialization.

%The GRU proposed in \cite{cho:14a} uses a gating mechanism with reset gates $r_t$ and update gates $z_t$ to capture dependencies within sequences. These gate control the composition of the new hidden state through linear interpolation and reduce the problem of vanishing gradients. For details about this unit, we refer the reader to \cite{cho:14a} We found in preliminary experiments that GRUs converged slightly faster than Long Short-Term Memory (LSTM) \cite{hochreiter:97a} for answer selection while achieving similar performance.
% At time $t$, the hidden state is defined as follows:$$h_t = (1-z_t)\odot h_{t-1} + z_t \odot \tilde{h}$$
%We implement two-way and one-way attention models as baselines.

%Each word $w_i^q$ in the question $q$, and word $w^a_l$ in the answer answer of length $L$ is embedded by an embedding matrix. We use pretrained Google News embeddings of dimensionality $d$, or 300 in this case. Let $e(w)$ return a vector $v \in \mathbb{R}^d$ from an embedding matrix $E$ corresponding to $w$ initalized from either the pretrained embeddings if it exists or otherwise a random vector sampled from the uniform distribution in the interval $[-0.25,0.25]$.
%, or $g(e(w_m)) = g_{forward}(e(w^q_m)) || g_{backward}(e(w^q_m))$ where $||$ is the vector concatenation operator.
%We implement a Attentive Pooling biGRU for answer sentence selection, which is a variation on the AP-biLSTM model \cite{santos:16a} that replaces the bidirectional Long Short-Term Memory Units (BiLSTM) with bidirectional Gated Recurrent Units (BiGRU).

\subsubsection*{Attentive Pooling}
\label{sssec:attentive-pooling}

Attentive Pooling (AP) is a framework-independent two-way attention mechanism that jointly learns a similarity measure between $q$ and $a$.
AP learns the similarity measure over the hidden states of $q$ and $a$. %When applied to recurrent neural networks, 
The AP matrix $H \in \mathbb{R}^{|q| \times |a|}$ has a bilinear form and is followed by a hyperbolic tangent non-linearity, where $U \in \mathbb{R}^{c \times c}$:
$$
H = \tanh(Q^T U A)
$$
The importance vectors $h^q \in \mathbb{R}^{|q|}$ and $h^a \in \mathbb{R}^{|a|}$ are generated from the column-wise and row-wise max pooling over $H$, respectively:
$$
[h^q]_j = \max\limits_{i \in [1,|q|]}[H_{j,i}]
$$
The normalized attention vectors $\sigma^q$ and $\sigma^a$ are created by applying the softmax activation function on $h^q$ and $h^a$:
$$
\sigma^q = \frac{\exp([h^q]_j)}{\sum\limits_{i \in [1,|q|]} \exp([h^q]_i)}
$$
The final representations $r^q = Q\sigma^q$ and $r^a = A\sigma^a$ for $q$ and $a$ are created using the dot products of the sentence representations and their corresponding attention vectors.
The score is computed for each pair using cosine similarity:
$$
s(q,a) = \frac{{r^q}^T r^a}{\lVert r^q \rVert \lVert r^a \rVert}
$$

\subsubsection*{One-Way Attention}
\label{sssec:one-way-attention}

Our one-way attention model is a simplified version of the attentive pooling model above, which is most similar to the global attention model introduced by \cite{luong:15a}.
We did not use the one-way attention from \cite{tan:15a} to avoid deviating the attention mechanism significantly.
Replacing $Q$ with $Q_{|q|}$, the last hidden state of $g$, $H$ becomes the importance vector $h$.
Again, we create the normalized attention vector $\sigma^a$ by applying the softmax activation function. The final representations are $r^q = Q_{|q|}$ and $r^a = A\sigma^a$.

%We construct each training example using a question $q$, a correct answer at index 0 of $q \in \mathbb{R}^n$, and $n-1$ incorrect answers in the other indices of $q$, where $n=50$.
%The training objective is defined as:
%\begin{eqnarray*}
%  L &=& \max\{0,m-s_\theta(q, a_0)+s_\theta(q,a_i)\}\\
%\end{eqnarray*}

\section{Experiments}
\label{sec:experiments}

\noindent Our systems are evaluated for the answer sentence selection (Section~\ref{ssec:answer-sentence-selection}) and answer triggering (Section~\ref{ssec:answer-triggering}) tasks on both WikiQA and our corpus.

%A thorough error analysis on each system with respect to our corpus is also provided.
%Section discusses experiments performed on both the WikiQA and our datasets. First, we briefly introduce the answer triggering task and propose our modified framework. Next, we switch to our data and present the results and error analysis. 
%The answer triggering task is significantly more difficult than a typical factoid question answering task due to the fact that the answer might not exist among the context.
%This dataset consists of questions retrieved from users' queries on the Bing search engine.
%For each question, a related Wikipedia abstract is collected and segmented into sentences.
%The number of candidate sentences per question varies between 1 and 30.

% =================================== SelQA ===================================

\subsection{SelQA: Selection-based QA Corpus}
\label{ssec:selqa}

\noindent Table~\ref{tbl:selqa} shows the distributions of our corpus, called SelQA.
Our corpus is split into training (70\%), development (10\%), and evaluation (20\%) sets.
The answer triggering data (AT) is significantly larger than the answer sentence selection data (ASS), due to the extra sections added by Task~5 (Section~\ref{ssec:annotation}).

\begin{table}[htbp!]
\caption{Distributions of our corpus. Q/Sec/Sen: number of questions/sections/sentences.}
\label{tbl:selqa}
\centering
\resizebox{\columnwidth}{!}{
\begin{tabular}{c||r||r|r||r|r}
 & & \multicolumn{2}{c||}{\bf ASS} & \multicolumn{2}{c}{\bf AT} \\
\cline{3-6}
\bf Set & \multicolumn{1}{c||}{\bf Q} & \multicolumn{1}{c|}{\bf Sec} & \multicolumn{1}{c||}{\bf Sen} & \multicolumn{1}{c|}{\bf Sec} & \multicolumn{1}{c}{\bf Sen} \\
\hline\hline
\texttt{TRN} & 5,529 & 5,529 & 66,438 & 27,645 & 205,075 \\
\texttt{DEV} &   785 &   785 &  9,377 & 3,925  &  28,798 \\
\texttt{TST} & 1,590 & 1,590 & 19,435 &  7,950 &  59,845 \\
\end{tabular}}
\end{table}

% =================================== Answer Sentence Selection ===================================

\subsection{Answer Sentence Selection}
\label{ssec:answer-sentence-selection}

\noindent Table~\ref{tbl:accuracies-wikiqa} shows results from ours and the previous approaches on WikiQA.
Two metrics are used, mean average precision (MAP) and mean reciprocal rank (MRR), for the evaluation of this task.
CNN$_0$ is our replication of the best model in \cite{yang:15a}.
CNN$_1$ and CNN$_2$ are the CNN models using the subtree matching in Section~\ref{ssec:cnn}, where the comparator of $f_c$ is either the word form or the word embedding respectively, and $f_m$ = \texttt{avg}.
The subtree matching models consistently outperforms the baseline model.
Note that among the three metrics of $f_m$, \texttt{avg}, \texttt{sum}, and \texttt{max}, \texttt{avg} outperformed the others in our experiments for the answer sentence selection task although no significant differences were found.
RNN$_0$ and RNN$_1$ are the RNN models using the one-way attention and the attentive pooling in Section~\ref{ssec:rnn}.
Note that RNN$_1$ converged much faster than RNN$_0$ at the same learning rate and fixed number of parameters in our experiments, implying that two-way attention assists with optimization.

\begin{table}[htp!]
\caption{Answer sentence selection results on the development and evaluation sets of WikiQA.}
\label{tbl:accuracies-wikiqa}
\centering
\resizebox{\columnwidth}{!}{
\begin{tabular}{l||c|c||c|c}
 & \multicolumn{2}{c||}{\bf Development} & \multicolumn{2}{c}{\bf Evaluation} \\
\cline{2-5}
\multicolumn{1}{c||}{\bf Model} & \bf MAP & \bf MRR & \bf MAP & \bf MRR \\
\hline\hline
CNN$_0$: baseline            &         69.93  &         70.66  &         65.62  &          66.46 \\
CNN$_1$: \texttt{avg} + word & \textbf{70.75} & \textbf{71.46} &         67.40  &          69.30 \\
CNN$_2$: \texttt{avg} + emb  &         69.22  &         70.18  & \textbf{68.78} & \textbf{70.82} \\
\hline
RNN$_0$: one-way             & \textbf{71.19} & \textbf{71.80} &         66.64  &         68.70  \\
RNN$_1$: attn-pool           &         67.56  &         68.31  & \textbf{67.47} & \textbf{68.92} \\
\hline\hline
Yang et al.~\cite{yang:15a}            & -              & -              &         65.20 &          66.52 \\
Santos et al.~\cite{santos:16a}         & -              & -              &         68.86 &          69.57 \\
Miao et al.~\cite{miao:15a}           & -              & -              &         68.86 &          70.69 \\
Yin et al.~\cite{yin:15a}            & -              & -              &         69.21 &          71.08 \\
Wang et al.~\cite{wang:16a}           & -              & -              &         70.58 &          72.26 \\
\end{tabular}}
\end{table}

\noindent It is interesting to see how CNN$_1$ and RNN$_0$ outperform CNN$_2$ and RNN$_1$ respectively on the development set, but not on the evaluation set.
This result may be explained by the larger percentage of overlapping words in the development set, enabling the simpler models perform more effectively.
%Without RNN$_0$, represents the entire query in its final hidden state, which can be difficult in longer queries.

\begin{table}[htp!]
\caption{Answer sent.\ selection results on SelQA.}
\label{tbl:accuracies-selqa}
\centering
\resizebox{\columnwidth}{!}{
\begin{tabular}{l||c|c||c|c}
 & \multicolumn{2}{c||}{\bf Development} & \multicolumn{2}{c}{\bf Evaluation} \\
\cline{2-5}
\multicolumn{1}{c||}{\bf Model} & \bf MAP & \bf MRR & \bf MAP & \bf MRR \\
\hline\hline
CNN$_0$: baseline            &         84.62  &         85.65  &         83.20  &         84.20  \\
CNN$_1$: \texttt{avg} + word &         85.04  &         86.17  &         84.00  &         84.94  \\
CNN$_2$: \texttt{avg} + emb  & \textbf{85.70} & \textbf{86.67} & \textbf{84.66} & \textbf{85.68} \\
\hline
RNN$_0$: one-way             &         82.26  &         83.68  &         82.06  &         83.18  \\
RNN$_1$: attn-pool           & \textbf{87.06} & \textbf{88.25} & \textbf{86.43} & \textbf{87.59} \\
\end{tabular}}
\end{table}

\noindent Table~\ref{tbl:accuracies-selqa} shows the results achieved by our models on SelQA.
CNN$_2$ outperforms the other CNN models, indicating the power of subtree matching coupled with word embeddings.
RNN$_1$ outperforms RNN$_0$, indicating the importance of attention over the questions.
Unlike the results on WikiQA in Table~\ref{tbl:accuracies-wikiqa}, CNN$_2$ and RNN$_0$ show the best performance on both the development and evaluation sets, implying the robustness of these models on our corpus.

\begin{table}[htp!]
\caption{MRR scores on the SelQA evaluation set for answer sentence selection with respect to topics.}
\label{tbl:accuracies-selqa-topic}
\centering
\resizebox{\columnwidth}{!}{
\begin{tabular}{l||c|c||c|c||c}
\bf Topic & \bf CNN$_0$ & \bf CNN$_2$ & \bf RNN$_0$ & \bf RNN$_1$ & \bf Q \\
\hline\hline
Arts             &         80.45  &         82.83  &         84.22  &         83.51  & 135 \\ 
Country          & \textbf{87.12} & \textbf{89.03} & \textbf{87.43} & \textbf{93.87} & 178\\ 
Food             &         85.30  &         86.11  &         84.72  &         86.74  & 147\\ 
H. Events        & \textbf{91.72} & \textbf{92.61} & \textbf{85.95} & \textbf{91.52} & 164\\ 
Movies           &         84.43  &         86.50  &         82.42  &         88.41  & 164\\ 
Music            &         81.38  &         80.39  &         84.57  &         84.38  & 155\\ 
Science          &         86.37  &         86.50  &         83.59  &         84.63  & 179\\ 
Sports           &         81.83  &         83.69  &         79.05  &         86.86  & 168\\ 
Travel           &         83.78  &         86.03  &         84.29  &         87.79  & 165 \\ 
TV               &         77.34  &         81.23  &         76.18  &         86.82  & 135\\ 
\end{tabular}}
\end{table}

\begin{figure*}[htbp!]
\centering
%\captionsetup{justification=centering}
\includegraphics[scale=0.23]{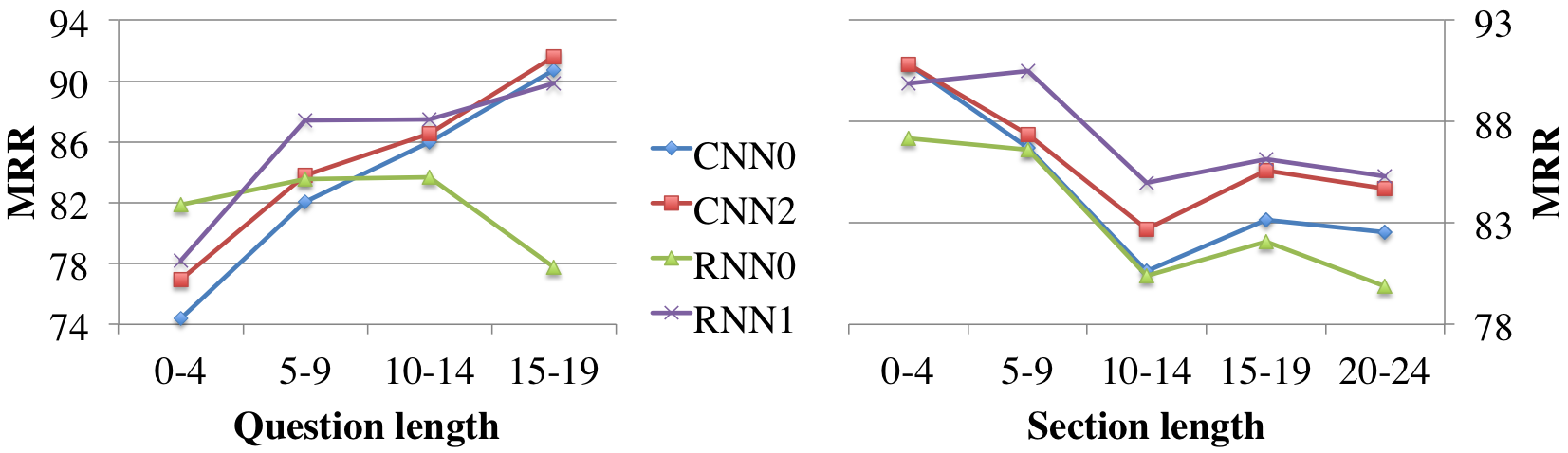}
\caption{Answer sentence selection on the SelQA evaluation set w.r.t.\ question and section lengths.}
\label{fig:length-ass}
\vspace{-2ex}
\end{figure*}

\noindent Table~\ref{tbl:accuracies-selqa-topic} shows the MRR scores from our models on SelQA with respect to different topics.
All models show strength on topics such as `Country' and `Historical Events', which is comprehensible since questions in these topics tend to be deterministic.
On the other hand, most models show weakness on topics such as `TV', `Arts', or `Music'.
This may be due to the fact that not many overlapping words are found between question and answer pairs in these documents, which also consist of many segments caused by bullet points.
%\noindent Tables~\ref{tbl:accuracies-selqa-topic},~\ref{tbl:accuracies-selqa-qtypes} and~\ref{tbl:accuracies-selqa-paraphrase} present MRR scores with respect to the topics of questions, their types, and whether the question is a paraphrase, respectively. 
%*$\downarrow$ and *$\uparrow$ are the baselines and the best performing approaches in Table~\ref{tbl:accuracies-selqa}, respectively.

\begin{table}[htp!]
\caption{MRR scores on the SelQA evaluation set for answer sentence selection w.r.t.\ paraphrasing.}
\label{tbl:accuracies-selqa-paraphrase}
\centering
\resizebox{\columnwidth}{!}{
\begin{tabular}{c||c|c||c|c||c}
\bf Type & \bf CNN$_0$ & \bf CNN$_2$ & \bf RNN$_0$ & \bf RNN$_1$ & \bf Q \\
\hline\hline
Original   &         86.70  &         88.31  &         85.57  &         89.90  & 810 \\ 
Paraphrase &         81.67  &         83.00  &         81.12  &         85.24  & 789 \\ 
\end{tabular}}
\end{table}

\noindent Table~\ref{tbl:accuracies-selqa-paraphrase} shows comparisons between questions from Tasks 1 and 2 (original) and Task 3 (paraphrase) in Section~\ref{ssec:annotation}.
As expected, noticeable performance drop is found for the paraphrased questions, which have much fewer overlapping words to the answer contexts than the original questions.

\begin{table}[htp!]
\caption{MRR scores on the SelQA evaluation set for answer sentence selection w.r.t.\ question types.}
\label{tbl:accuracies-selqa-qtypes}
\centering
\resizebox{\columnwidth}{!}{
\begin{tabular}{l||c|c||c|c||c}
\bf Type & \bf CNN$_0$ & \bf CNN$_2$ & \bf RNN$_0$ & \bf RNN$_1$ & \bf Q \\
\hline\hline
What  &         84.54  &         85.36  &         83.50  &         87.66  & 678 \\ 
How   &         81.92  &         84.01  &         82.04  &         87.32  & 233 \\ 
Who   & \textbf{85.46} & \textbf{88.17} &         80.36  &         85.99  & 195 \\ 
When  &         84.21  &         85.56  & \textbf{86.16} & \textbf{90.35} & 180 \\ 
Where &         83.78  &         87.44  &         84.59  &         82.54  &  85 \\ 
Why   &         78.55  &         82.64  &         80.61  &         84.07  &  41 \\ 
Misc. &         84.17  &         84.80  &         85.20  &         89.66  & 215 \\
\end{tabular}}
\end{table}

\noindent Table~\ref{tbl:accuracies-selqa-qtypes} shows the MRR scores with respect to question types.
The CNN models show strength on the `who' type, whereas the RNN models show strength on the `when' type.
Each model varies on showing their weakness, which we will explore in the future.
Finally, Figure~\ref{fig:length-ass} shows the performance difference with respect to question and section lengths.
All models except for RNN$_0$ tend to perform better as questions become longer.
This makes sense since longer questions are usually more informative.
On the other hand, models generally perform worse as sections become longer, which also makes sense because the models have to select the answer contexts from larger pools.

% =================================== Answer Triggering ===================================

\subsection{Answer Triggering}
\label{ssec:answer-triggering}

\noindent Due to the nature of answer triggering, metrics used for evaluating answer sentence selection are not used here, because those metrics assume that models are always provided with contexts including the answers.
Broadly speaking, the answer sentence selection task is a raking problem, while answer triggering is a binary classification task with additional constraints.
Thus, the F1-score on the question level was proposed by \cite{yang:15a} as the evaluation for this task, which we follow.

Table~\ref{tbl:at-accuracies} shows the answer triggering results on WikiQA.
Note that RNN$_0$ using one-way attention was dropped for these experiments because it did not show comparable performance against the others for this task.
Interestingly, the CNN model with $f_m$ = \texttt{max} outperformed the other metrics for answer triggering, although \texttt{avg} was found to be the most effective for answer sentence selection.
The CNN subtree matching models consistently gave over 2\% improvements to the baseline model.

\begin{table}[htp!]
\caption{Answer triggering results on WikiQA.}
\label{tbl:at-accuracies}
\centering
\resizebox{\columnwidth}{!}{
\begin{tabular}{l||c|c|c||c|c|c}
 & \multicolumn{3}{c||}{\bf Development} & \multicolumn{3}{c}{\bf Evaluation} \\
\cline{2-7}
\multicolumn{1}{c||}{\bf Model} & \bf P & \bf R & \bf F1 & \bf P & \bf R & \bf F1 \\
\hline\hline
CNN$_0$: baseline            & 41.86  &   42.86   &         42.35  & 29.70 &     37.45     &         32.73 \\
CNN$_1$: \texttt{max} + word & 44.53  &   45.24   & \textbf{44.88} & 29.77 &     42.39     &         34.97 \\
CNN$_2$: \texttt{max} + emb  & 43.07  &   46.83   &         44.87  & 29.77 &     42.39     &         34.97 \\
CNN$_3$: \texttt{max} + emb+ & 44.44  &   44.44   &         44.44  & 29.43 &     48.56     & \textbf{36.65} \\
\hline
RNN$_1$: attn-pool           & 25.95  &   38.10   &         30.87  & 24.32 &     47.74     &         32.22 \\
\hline\hline
Yang et al.~\cite{yang:15a}           & -      & -         & -              & 27.96 &     37.86     &         32.17 \\
\end{tabular}}
\end{table}

\begin{figure*}[htbp!]
\centering
%\captionsetup{justification=centering}
\includegraphics[scale=0.23]{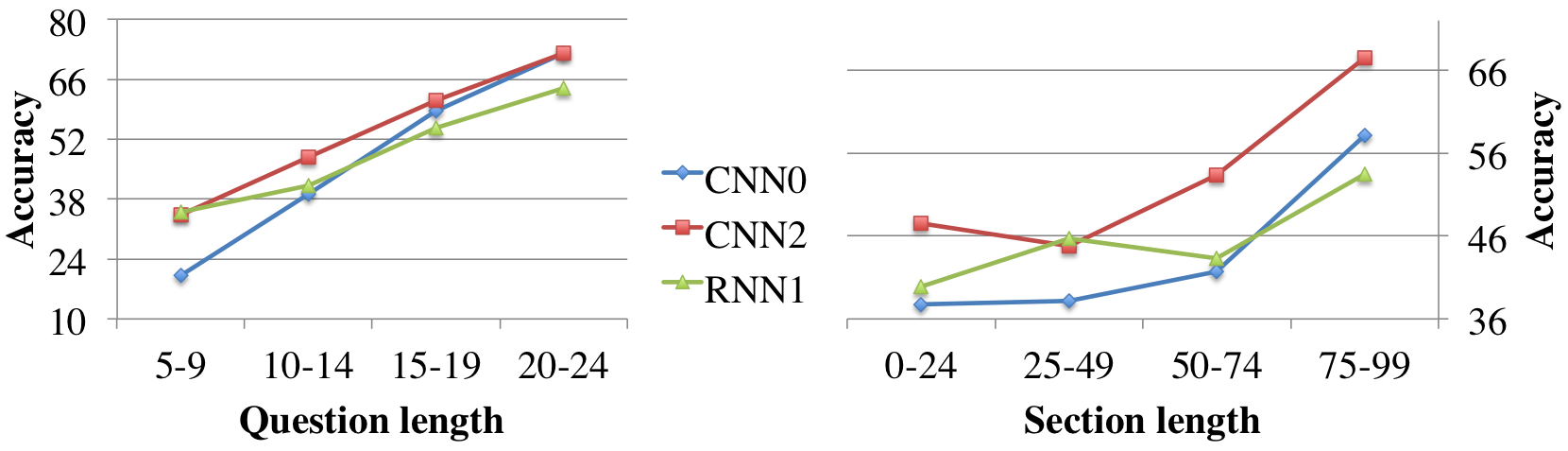}
\caption{Answer triggering on the SelQA evaluation set w.r.t.\ question and section lengths.}
\label{fig:length-at}
\vspace{-2ex}
\end{figure*}

\noindent In addition, CNN$_3$ was experimented by retraining word embeddings (emb+), which performed slightly worse on the development set, but gave another 1.68\% improvement on the evaluation set.\footnote{Retraining word embeddings was not found to be useful for answer sentence selection.}
RNN$_1$ showed a very similar result to \cite{yang:15a}, which was surprising since it performed so much better for answer sentence selection.
This can be due to a lack of hyper-parameter optimization, which we leave as a future work.

\begin{table}[htp!]
\caption{Answer triggering results on SelQA.}
\label{tbl:at-accuracies-selqa}
\centering
\resizebox{\columnwidth}{!}{
\begin{tabular}{l||c|c|c||c|c|c}
 & \multicolumn{3}{c||}{\bf Development} & \multicolumn{3}{c}{\bf Evaluation} \\
\cline{2-7}
\multicolumn{1}{c||}{\bf Model} & \bf P & \bf R & \bf F1 & \bf P & \bf R & \bf F1 \\
\hline\hline
CNN$_0$: baseline            &   50.63   &  40.60  &  45.07  &  52.10  &  40.34  &  45.47   \\
CNN$_1$: \texttt{max} + word &   48.15   &  47.99  &  48.07  &  52.22  &  47.30  &  49.64   \\
CNN$_2$: \texttt{max} + emb  &   49.32   &  48.99  &  \textbf{49.16}  &  53.69  &  48.38  &  \textbf{50.89}   \\
CNN$_3$: \texttt{max} + emb+ &   47.16�  &��47.32  &��47.24  &  52.14  &��47.14  &� 49.51   \\
\hline
RNN$_1$: attn-pool           &   45.52   &  42.62  & 44.02   &  47.96  &  43.59  & 45.67    \\
\end{tabular}}
\end{table}

\noindent Table~\ref{tbl:at-accuracies-selqa} shows the answer triggering results on SelQA.
Unlike the results on WikiQA (Table~\ref{tbl:at-accuracies}), CNN$_2$ outperforms CNN$_3$ on our corpus.
On the other hand, RNN$_1$ shows a similar score to \cite{yang:15a} as it does on WikiQA.
CNN$_2$ using subtree matching gives over a 5\% improvement to the baseline model, which is significant.

Table~\ref{tbl:at-accuracies-selqa-topic} shows the accuracies on SelQA with respect to different topics.
The accuracy is measured on the subset of questions that contain at least one answer among candidates; the top ranked sentence is taken and checked for the correct answer.
Similar to answer sentence selection, CNN$_2$ stills shows strength on topics such as `Country' and `Historical Events', but the trend is not as clear for the other models. The worst performing topics are `TV', `Music' and `Art'. Such a noticeable difference might be caused by the unusual semantic sentence constructions of the text. Sections in these categories often contain listings, bullet-pointed texts etc., which is problematic for the models to properly take care of. How to correctly understand and solve question from such context will be a challenge to the future systems.
Also, interestingly, the standard deviation is much smaller for RNN$_1$ (3.9\%) compared to the CNN models (10-12\%) although RNN$_1$'s overall performance is lower.

\begin{table}[htp!]
\caption{Accuracies on the SelQA evaluation set for answer triggering with respect to topics.}
\label{tbl:at-accuracies-selqa-topic}
\centering
%\resizebox{\columnwidth}{!}{
\begin{tabular}{l||c|c||c||c}
\bf Topic & \bf CNN$_0$ & \bf CNN$_2$ & \bf RNN$_1$ & \bf Q \\
\hline\hline
Arts              &         27.45  &         31.37  &         43.14  & 135 \\
Country           &         43.59  & \textbf{61.54} &         38.46  & 178\\
Food              &         31.40  &         44.19  &         46.51  & 147\\
H. Events         & \textbf{60.32} & \textbf{63.49} &         38.10  & 164\\
Movies            &         37.74  &         45.28  &         39.62  & 164\\
Music             &         29.31  &         36.21  &         44.83  & 155\\
Science           &         45.00  &         57.50  &         43.75  & 179\\
Sports            &         50.00  &         58.11  &         47.30  & 168\\
Travel            &         42.68  &         50.00  & \textbf{48.78} & 165 \\
TV                &         32.79  &         32.79  &         39.34  & 135\\
\end{tabular}%}
\end{table}

\noindent Table~\ref{tbl:at-accuracies-selqa-paraphrase} shows the accuracies on SelQA with respect to paraphrasing, which is similar to the trend found in Table~\ref{tbl:accuracies-selqa-paraphrase} for answer sentence selection.

\begin{table}[htp!]
\caption{Accuracies on the SelQA evaluation set for answer triggering w.r.t.\ paraphrasing.}
\label{tbl:at-accuracies-selqa-paraphrase}
\centering
%\resizebox{\columnwidth}{!}{
\begin{tabular}{c||c|c||c||c}
\bf Type & \bf CNN$_0$ & \bf CNN$_2$ & \bf RNN$_1$ & \bf Q \\
\hline\hline
Original   & 46.15 & 55.13 &  44.36 & 810 \\
Paraphrase & 31.52 & 38.52 &  42.21 & 789 \\
\end{tabular}%}
\end{table}

\noindent Table~\ref{tbl:at-accuracies-selqa-qtypes} shows the accuracies on SelQA with respect to question types.
Interestingly, each model shows different strength on different types, which may suggest a possibility of an ensemble model.
Finally, Figure~\ref{fig:length-at} shows the performance difference with respect to question and section lengths for the answer triggering task. All the models tend to perform better as questions become longer. Similarly as in the answer sentence selection task, since longer questions are more informative, it is understandable.
Interestingly, once the section becomes longer, the accuracy increases. We hypothesize that such a behavior might be caused by the fact that
it is easier for the models to decide whether the context of the section is the same as the context of the question when there is more information (sentences) in the section. Thus, this phenomenon is related to the task of answer triggering, where the model not only choose the sentence with the answer, but must decide if the context matches first.

\begin{table}[htp!]
\caption{Accuracies on the SelQA evaluation set for answer triggering w.r.t.\ question types.}
\label{tbl:at-accuracies-selqa-qtypes}
\centering
%\resizebox{\columnwidth}{!}{
\begin{tabular}{l||c|c||c||c}
\bf Type & \bf CNN$_0$ & \bf CNN$_2$ & \bf RNN$_1$ & \bf Q \\
\hline\hline
What  &         40.68  &         50.19  &         44.11  & 678 \\
How   &         36.63  &         43.56  &         44.55  & 233 \\
Who   & \textbf{44.94} &         50.56  &         38.20  & 195 \\
When  &         33.33  &         43.06  &         38.89  & 180 \\
Where &         33.33  & \textbf{51.85} &         40.74  &  85 \\
Why   &         42.11  &         47.37  & \textbf{57.89} &  41 \\
Misc. &         44.90  &         51.02  &         46.94  & 215 \\
\end{tabular}%}
\end{table}

\section{Conclusion}
\label{sec:conclusion}

\noindent In this paper we present a new benchmark for two major question answering tasks: answer sentence selection and answer triggering.
Several systems using neural networks are developed for the analysis of our corpus.
Our analysis shows different aspects about the current QA approaches, beneficial for further enhancement.

Researchers devoted to relatively small datasets reveal useful characteristics of the question answering tasks.
Techniques that result in improvements on smaller datasets are often significantly diminished with larger datasets.
Current hardware trends and the availability of larger datasets make large scale question answering more accessible.

We plan to continue our work on providing large scale corpora for open-domain question answering. Also, we intend to continue working towards providing context-aware frameworks for question answering. %Currently existing state-of-the-art systems are not capable of solving problems that are semantically complicated.

%Next, we acknowledge already existing approach based on convolutional neural network and extend it with matching syntactic knowledge in dependency structured data. The incremental scores are reported for two publicly available data sets for answer sentence selection and answer triggering: WikiQA and new corpus described in this paper. Finally, our evaluation shows that dependency matching techniques as well as the addition of the embedding layer provide a significant boost of over 4\% to the current state-of-the-art.

%However, overfitting limits the analysis of these tasks since we gain significant generalization performance on larger datasets.

\section*{Acknowledgement}
We gratefully acknowledge the support from Infosys Ltd.  Any contents in this material are those of the authors and do not necessarily reflect the views of Infosys Ltd.

\bibliography{ICTAI-2016}

% Generated by IEEEtran.bst, version: 1.14 (2015/08/26)
\begin{thebibliography}{10}
\providecommand{\url}[1]{#1}
\csname url@samestyle\endcsname
\providecommand{\newblock}{\relax}
\providecommand{\bibinfo}[2]{#2}
\providecommand{\BIBentrySTDinterwordspacing}{\spaceskip=0pt\relax}
\providecommand{\BIBentryALTinterwordstretchfactor}{4}
\providecommand{\BIBentryALTinterwordspacing}{\spaceskip=\fontdimen2\font plus
\BIBentryALTinterwordstretchfactor\fontdimen3\font minus
  \fontdimen4\font\relax}
\providecommand{\BIBforeignlanguage}[2]{{%
\expandafter\ifx\csname l@#1\endcsname\relax
\typeout{** WARNING: IEEEtran.bst: No hyphenation pattern has been}%
\typeout{** loaded for the language `#1'. Using the pattern for}%
\typeout{** the default language instead.}%
\else
\language=\csname l@#1\endcsname
\fi
#2}}
\providecommand{\BIBdecl}{\relax}
\BIBdecl

\bibitem{wang:07a}
M.~Wang, N.~A. Smith, and T.~Mitamura, ``{What is the Jeopardy Model? A
  Quasi-Synchronous Grammar for QA},'' in \emph{Proceedings of the Joint
  Conference on Empirical Methods in Natural Language Processing and
  Computational Natural Language Learning}, ser. EMNLP-CoNLL'07, 2007, pp.
  22--32.

\bibitem{yang:15a}
Y.~Yang, W.-t. Yih, and C.~Meek, ``{WIKIQA: A Challenge Dataset for Open-Domain
  Question Answering},'' in \emph{Proceedings of the Conference on Empirical
  Methods in Natural Language Processing}, ser. EMNLP'15, 2015, pp. 2013--2018.

\bibitem{feng:15a}
M.~Feng, B.~Xiang, M.~R. Glass, L.~Wang, and B.~Zhou, ``{Applying Deep Learning
  to Answer Selection: A Study and An Open Task},'' in \emph{IEEE Workshop on
  Automatic Speech Recognition and Understanding}, 2015, pp. 813--820.

\bibitem{yu:14a}
L.~Yu, K.~M. Hermann, P.~Blunsom, and S.~Pulman, ``{Deep Learning for Answer
  Sentence Selection},'' in \emph{Proceedings of the NIPS Deep Learning
  Workshop}, 2014.

\bibitem{wang:15a}
D.~Wang and E.~Nyberg, ``{A Long Short-Term Memory Model for Answer Sentence
  Selection in Question Answering},'' in \emph{Proceedings of the 53rd Annual
  Meeting of the Association for Computational Linguistics and the 7th
  International Joint Conference on Natural Language Processing}, ser. ACL'15,
  2015, pp. 707--712.

\bibitem{severyn:15a}
A.~Severyn and A.~Moschitti, ``{Learning to Rank Short Text Pairs with
  Convolutional Deep Neural Networks},'' in \emph{Proceedings of the 38th
  International ACM SIGIR Conference on Research and Development in Information
  Retrieval}, ser. SIGIR '15, 2015, pp. 373--382.

\bibitem{yu2014deep}
L.~Yu, K.~M. Hermann, P.~Blunsom, and S.~Pulman, ``Deep learning for answer
  sentence selection,'' \emph{arXiv preprint arXiv:1412.1632}, 2014.

\bibitem{iyyer2014neural}
M.~Iyyer, J.~Boyd-Graber, L.~Claudino, R.~Socher, and H.~Daum{\'e}~III, ``A
  neural network for factoid question answering over paragraphs,'' in
  \emph{Proceedings of the 2014 Conference on Empirical Methods in Natural
  Language Processing (EMNLP)}, 2014, pp. 633--644.

\bibitem{dong2015question}
L.~Dong, F.~Wei, M.~Zhou, and K.~Xu, ``Question answering over freebase with
  multi-column convolutional neural networks,'' in \emph{Proceedings of the
  53rd Annual Meeting of the Association for Computational Linguistics and the
  7th International Joint Conference on Natural Language Processing}, vol.~1,
  2015, pp. 260--269.

\bibitem{yin2015abcnn}
W.~Yin, H.~Sch{\"u}tze, B.~Xiang, and B.~Zhou, ``Abcnn: Attention-based
  convolutional neural network for modeling sentence pairs,'' \emph{arXiv
  preprint arXiv:1512.05193}, 2015.

\bibitem{yih2014semantic}
W.-t. Yih, X.~He, and C.~Meek, ``Semantic parsing for single-relation question
  answering,'' in \emph{Proceedings of ACL}, 2014.

\bibitem{blunsom2014convolutional}
P.~Blunsom, E.~Grefenstette, N.~Kalchbrenner \emph{et~al.}, ``A convolutional
  neural network for modelling sentences,'' in \emph{Proceedings of the 52nd
  Annual Meeting of the Association for Computational Linguistics}.\hskip 1em
  plus 0.5em minus 0.4em\relax Proceedings of the 52nd Annual Meeting of the
  Association for Computational Linguistics, 2014.

\bibitem{heilman:10a}
M.~Heilman and N.~A. Smith, ``{Tree Edit Models for Recognizing Textual
  Entailments, Paraphrases, and Answers to Questions},'' in \emph{Human
  Language Technologies: The 2010 Annual Conference of the North American
  Chapter of the Association for Computational Linguistics}, ser. HLT'10, 2010,
  pp. 1011--1019.

\bibitem{wang:10a}
M.~Wang and C.~Manning, ``{Probabilistic Tree-Edit Models with Structured
  Latent Variables for Textual Entailment and Question Answering},'' in
  \emph{Proceedings of the 23rd International Conference on Computational
  Linguistics}, ser. COLING'10, 2010, pp. 1164--1172.

\bibitem{severyn:13a}
A.~Severyn and A.~Moschitti, ``{Automatic Feature Engineering for Answer
  Selection and Extraction},'' in \emph{Proceedings of the Conference on
  Empirical Methods in Natural Language Processing}, ser. EMNLP'13, 2013, pp.
  458--467.

\bibitem{tan:15a}
M.~Tan, B.~Xiang, and B.~Zhou, ``{LSTM-based Deep Learning Models for
  Non-factoid Answer Selection},'' \emph{arXiv}, vol. arXiv:1511.04108, 2015.

\bibitem{santos:16a}
\BIBentryALTinterwordspacing
C.~N.~d. Santos, M.~Tan, B.~Xiang, and B.~Zhou, ``Attentive pooling networks,''
  \emph{CoRR}, vol. abs/1602.03609, 2016. [Online]. Available:
  \url{http://arxiv.org/abs/1602.03609}
\BIBentrySTDinterwordspacing

\bibitem{ho:15a}
C.-J. Ho, A.~Slivkins, S.~Suri, and J.~W. Vaughan, ``{Incentivize High Quality
  Crowdwork},'' in \emph{Proceedings of the 24th World Wide Web Conference},
  ser. WWW'15, 2015.

\bibitem{mikolov:13b}
T.~Mikolov, I.~Sutskever, K.~Chen, G.~S. Corrado, and J.~Dean, ``{Distributed
  Representations of Words and Phrases and their Compositionality},'' in
  \emph{Proceedings of Advances in Neural Information Processing Systems 26},
  ser. NIPS'13, 2013, pp. 3111--3119.

\bibitem{choi:13a}
J.~D. Choi and A.~McCallum, ``{Transition-based Dependency Parsing with
  Selectional Branching},'' in \emph{Proceedings of the 51st Annual Meeting of
  the Association for Computational Linguistics}, ser. ACL'13, 2013, pp.
  1052--1062.

\bibitem{cho:14a}
K.~Cho, B.~van Merrienboer, C.~Gulcehre, D.~Bahdanau, F.~Bougares, H.~Schwenk,
  and Y.~Bengio, ``{Learning Phrase Representations using RNN Encoder--Decoder
  for Statistical Machine Translation},'' in \emph{Proceedings of the
  Conference on Empirical Methods in Natural Language Processing}, ser.
  EMNLP'14, 2014, pp. 1724--1734.

\bibitem{luong:15a}
T.~Luong, H.~Pham, and C.~D. Manning, ``{Effective Approaches to
  Attention-based Neural Machine Translation},'' in \emph{Proceedings of the
  2015 Conference on Empirical Methods in Natural Language Processing}, ser.
  EMNLP'15, 2015, pp. 1412--1421.

\bibitem{miao:15a}
Y.~Miao, L.~Yu, and P.~Blunsom, ``{Neural Variational Inference for Text
  Processing},'' \emph{arXiv}, vol. arXiv:1511.06038, 2015.

\bibitem{yin:15a}
W.~Yin, H.~Sch{\"u}tze, B.~Xiang, and B.~Zhou, ``{ABCNN: Attention-Based
  Convolutional Neural Network for Modeling Sentence Pairs},'' \emph{arXiv},
  vol. arXiv:1512.05193, 2015.

\bibitem{wang:16a}
Z.~Wang, H.~Mi, and A.~Ittycheriah, ``{Sentence Similarity Learning by Lexical
  Decomposition and Composition},'' \emph{arXiv}, vol. arXiv:1602.07019, 2016.

\end{thebibliography}
\bibliographystyle{IEEEtran}

\end{document}